\documentclass{article}

\usepackage{times}
\usepackage{soul}
\usepackage{url}
\usepackage{hyperref}
\usepackage[small]{caption}
\usepackage{amsmath}
\usepackage{amsthm}
\usepackage[switch]{lineno}
\usepackage{longtable}
\usepackage{supertabular}
\usepackage{multirow}

\usepackage{tikz}
\usepackage[edges]{forest}

\usepackage[in]{fullpage}  
\usepackage{arxiv} 
\usepackage{mathpazo} 

\usepackage[utf8]{inputenc} 
\usepackage[T1]{fontenc}    
\usepackage{url}            
\usepackage{booktabs}       
\usepackage{amsfonts}       
\usepackage{microtype}      
\usepackage{lipsum}         

\usepackage{graphicx}  
\usepackage{float} 
\usepackage{subfigure} 
\usepackage{algorithm}
\usepackage{algorithmic}
\usepackage{tabulary}

\usepackage{amssymb}
\usepackage{pifont}

\newcommand{\methodname}{LS-MBPO}

\usepackage{algorithm}
\usepackage{algorithmic}

\usepackage{adjustbox}
\usepackage{capt-of}

\hypersetup{
    colorlinks=true,
    citecolor=blue,
    linkcolor=blue,
    citecolor=blue
}

\title{Efficient VLA Reinforcement Learning via Subchunk Rollouts in Latent World Models}

\usepackage{authblk}
\author{%
  Nan Tang\textsuperscript{\rm 1,2,*},
  Han-Tao Zhang\textsuperscript{\rm 1,*},
  Jing-Cheng Pang\textsuperscript{\rm 1,*},
  Hao-Nan Wang\textsuperscript{\rm 1,2,*}, 
  Hao-Xiang Ren\textsuperscript{\rm 1,2}, \\ \vspace{-1.0em}
  Yi-Fei Sheng\textsuperscript{\rm 1,2}, 
  Jun-Yin Ye\textsuperscript{\rm 2},
  Tian Xu\textsuperscript{\rm 1}, 
  Fan-Ming Luo\textsuperscript{\rm 1,2},
  Hao-Xin Lin\textsuperscript{\rm 1,2},
  Yang Yu\textsuperscript{\rm 1,$\diamond$}\\
  \textsuperscript{\rm 1} National Key Laboratory for Novel Software Technology \& \\
School of Artificial Intelligence, Nanjing University, China \\
 \textsuperscript{\rm 2}Cirquar Technologies, Nanjing, China \\
  \textsuperscript{*} Equal contribution\\
  \textsuperscript{$\diamond$} Corresponding: yuy@nju.edu.cn
}

\date{}
\begin{document}

\maketitle

\begin{abstract}

Model-based reinforcement learning (MBRL) can improve pretrained vision-language-action (VLA) policies through imagined experiences. However, existing VLA-MBRL methods primarily rely on large-scale image or video generation world models, incurring large computational overhead. 
Latent world models offer a more efficient alternative but face two challenges: compounding errors during recursive prediction and incompatibility with VLAs' visual input interfaces.
In this work, we propose LS-MBPO, a lightweight residual MBRL framework that leverages latent world models to improve upon pretrained VLAs. 
LS-MBPO introduces two key techniques:
(1) Hierarchical Forcing, which combines dense causal supervision with autoregressive feedback to mitigate compounding prediction errors;
and (2) Latent Subchunk Rollouts, which leverage action chunks from a frozen VLA policy and a latent residual policy for successive subchunk refinement, bridging the gap between latent imagination and VLA policy execution.
We further introduce a Progress-Conditioned Branch-Start Curriculum that progressively shifts rollout starting points from later to earlier task stages, enabling effective policy learning from short imagined rollouts. 
Extensive experiments on LIBERO, RoboTwin, and a real-world ALOHA robot show that LS-MBPO achieves superior task performance and training efficiency compared to existing methods.

\end{abstract}    
\section{Introduction}
\label{sec:intro}

World models~\cite{howshouldworldmodels,vlambpo} provide a promising avenue for improving vision-language-action (VLA) policies through model-based reinforcement learning (MBRL)~\cite{MBPO}, as they enable the generation of imagined experiences for policy optimization while reducing reliance on costly real-world interactions. Recent approaches~\cite{vlambpo,WMPO,UGROW} have primarily explored this direction by leveraging large-scale generative world models to predict future observations in the pixel space. However, generating high-dimensional visual observations introduces substantial computational overhead, while pixel-level prediction does not necessarily provide the compact state representations most suitable for downstream policy learning. Latent world models offer a promising alternative by modeling environment dynamics directly in a compact representation space, avoiding the need for explicit future image generation. Moreover, recent advances in latent world modeling, such as LeWorldModel~\cite{LeWM}, demonstrate the potential of learning task-relevant representations that capture essential information for downstream decision-making. These properties make latent world models particularly appealing for developing more computationally efficient MBRL frameworks for VLA policies.

\begin{figure*}[t]
    \centering
    \vspace{0em}
    \includegraphics[width=0.75\linewidth]{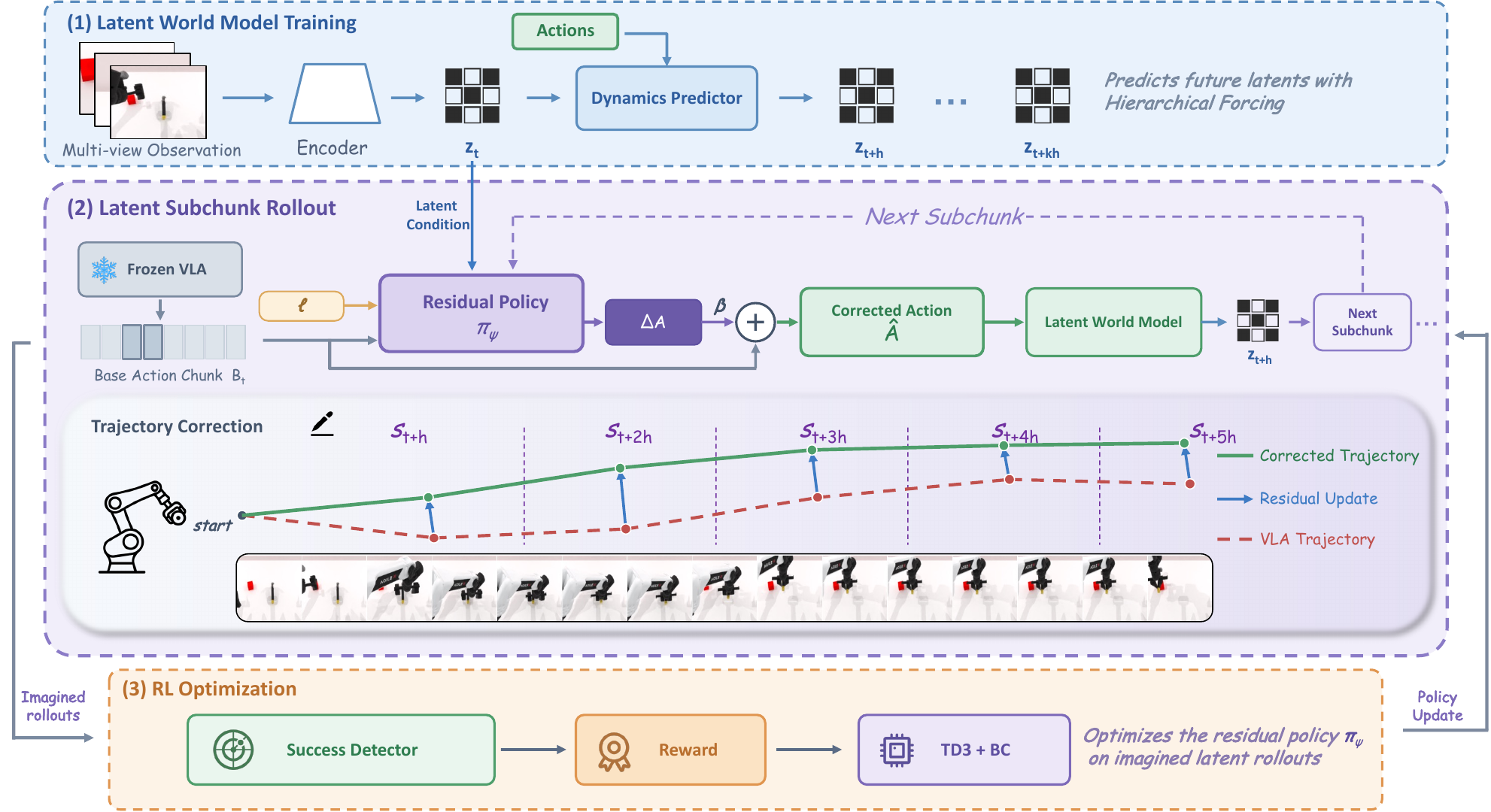}
    \vspace{0.5em}
    \caption{Overview of \methodname. (1) \methodname~first trains a latent world model to predict future latents with Hierarchical Forcing. (2) \methodname~uses Latent Subchunk Rollout to generate imagined experiences with a residual policy attached to a frozen VLA. (3) A success detector is trained to reward the imagined rollouts, with which a residual policy is optimized using TD3 + BC.
    }
    \label{fig:method_overview}
    \vspace{-1.5em}
\end{figure*}

Despite these advantages, integrating latent world models into VLA-based MBRL remains challenging due to two fundamental challenges. First, latent world models suffer from compounding prediction errors~\cite{talvitie2014model,MBPO,xu2020} during recursive rollouts. Existing approaches~\cite{zhou2025dinowm,LeWM} typically train latent dynamics using one-step prediction objectives, where predictions are conditioned on ground-truth latent histories. However, during imagined rollouts, the model must recursively take its own predictions as inputs, creating a discrepancy between training and inference conditions. Prediction errors at earlier steps can propagate to subsequent predictions, while the resulting deviations in latent inputs may further amplify these errors, leading to increasingly unreliable imagined trajectories. Second, there exists a mismatch between the interfaces of latent world models and pretrained VLAs. Latent world models predict future states in compact latent spaces, whereas pretrained VLAs typically require visual observations as inputs, making it difficult to directly use predicted latent states for subsequent action generation. Besides the interface mismatch, existing VLA-MBRL approaches, such as VLA-MBPO~\cite{vlambpo}, often rely on fine-tuning all VLA parameters using imagined experiences. Such full-parameter optimization introduces substantial computational and memory overhead, which undermines the computational advantage of using latent world models for policy learning. 
These challenges raise an important question: how can we leverage latent imagination for VLA reinforcement learning while ensuring reliable recursive prediction and computationally efficient policy optimization?

To address these challenges, we propose \underline{L}atent \underline{S}ubchunk Rollout for \underline{MBPO} (\methodname), a lightweight residual model-based reinforcement learning framework for VLA using latent world models, which is shown in Fig.~\ref{fig:method_overview}. 
First, we introduce \textit{Hierarchical Forcing} to improve the reliability of latent dynamics under recursive prediction. It combines dense causal prediction supervision within each temporal window with autoregressive feedback across consecutive windows, enabling the world model to progressively adapt to its own predictions while mitigating compounding errors. 
Second, we propose \textit{Latent Subchunk Rollouts} to bridge the interface gap between pretrained VLAs and latent world models. Specifically, we freeze the pretrained VLA and divide its generated action chunk into shorter subchunks as base actions, while training a lightweight residual policy conditioned on the current latent state and the corresponding base actions. During imagination, the residual policy progressively refines the base actions based on the updated latent states predicted by the world model. In this way, a single VLA query provides a behavioral prior for multiple successive residual decisions, allowing the policy to be optimized through the lightweight residual policy and value network.
Furthermore, we introduce a \textit{Progress-Conditioned Branch-Start Curriculum} that gradually shifts the sampling distribution from states near task completion to earlier task stages, enabling more efficient learning from imagined rollouts.
On LIBERO Long, \methodname~completes the world model training on 4 RTX 5090s and the RL stage on a single
RTX 5090 with a success rate of 71.2\%, compared with
64.4\% for VLA-MBPO trained on 4 H100 GPUs, demonstrating a
practical route towards resource-efficient VLA policy improvement.

Our contributions are as follows:
First, we propose \methodname, a lightweight residual model-based reinforcement learning framework for improving VLA with latent world models. To our knowledge, this is the first work attempting to integrate VLA with latent world models for model-based RL.
Second, we address two core limitations of VLA-MBRL in latent world models, i.e., compounding error and interface mismatch by proposing Hierarchical Forcing and Latent Subchunk Rollouts.
Finally, through extensive experiments on LIBERO, RoboTwin and a real-world ALOHA robot, we demonstrate that \methodname~not only improves sample efficiency but also achieves superior policy performance, offering a practical solution to efficient VLA reinforcement learning through latent imagination.
\section{Related Works}
\label{sec:related_works}

\subsection{World Models for Visual Control}
World models have been widely explored for visual control by learning action-conditioned environment dynamics. Early approaches, such as Visual Foresight \cite{finn2017deep}, predict future visual observations to support planning. Advances in generative modeling have enabled increasingly general and realistic world simulation: UniSim \cite{yang2023learning} learns an interactive simulator supporting vision-language planning and reinforcement learning, while NVIDIA Cosmos \cite{cosmos2025} demonstrates the potential of large-scale video models for physical world simulation and synthetic data generation. More recently, Ctrl-World \cite{guo2025ctrl} enables policy-in-the-loop imagination for robot manipulation, allowing pretrained vision-language-action (VLA) policies to be evaluated and improved using generated visual trajectories. However, synthesizing high-dimensional visual observations incurs substantial computational overhead, and visual fidelity alone does not guarantee representations suited to downstream control. In parallel, latent world models learn action-conditioned dynamics in compact representation spaces, supporting planning and policy learning without decoding images during imagined rollouts. Representative methods, including PlaNet \cite{hafner2019learning}, Dreamer \cite{hafner2020dreamer}, and TD-MPC \cite{hansen2022temporal}, demonstrate the effectiveness of latent-space imagination for model-based control. More recent methods \cite{reviwo,zhou2025dinowm,LeWM,sg_jepa,h_jepa} further explore visual control through predictive dynamics over pretrained visual features and jointly learned representations. These advances motivate investigating how latent-space imagination can support efficient reinforcement learning of VLA policies while remaining compatible with their visual observation interfaces.

\subsection{Reinforcement Learning for VLA Models}
Recent works have explored reinforcement learning to improve pretrained VLA policies. Model-free approaches, such as $\pi_{\mathrm{RL}}$ \cite{pi_rl}, SimpleVLA-RL \cite{simplevla_rl}, and $\pi^*_{0.6}$ \cite{pi06}, optimize VLA policies through environment interactions. RLT \cite{rlt} introduces a lightweight actor-critic architecture to refine the actions of a frozen pretrained policy, avoiding full-parameter policy updates.
Model-based reinforcement learning further enables VLA policies to learn from imagined experiences generated by world models. WMPO \cite{WMPO} leverages generative world models to perform policy optimization through imagined rollouts. VLA-MBPO \cite{vlambpo} introduces multi-view world modeling and branched rollouts for model-based VLA fine-tuning. Later, U-GROW \cite{UGROW} improves rollout efficiency through uncertainty-guided starting state selection. 
Despite these advances, existing VLA-MBRL methods primarily rely on pixel-based world models, requiring the generation of high-dimensional visual observations during imagination. Although latent world models offer a more computationally efficient alternative, their integration with pretrained VLAs remains challenging due to the mismatch between predicted latent states and the visual inputs expected by VLAs. Our work addresses this challenge by combining latent-space subchunk imagination with efficient residual policy learning.
\section{Method}
This section presents the method \methodname, a lightweight residual model-based reinforcement learning framework for VLA using latent world models. It is supposed that there exists an offline dataset $\mathcal{D}$ and a small-scale expert dataset $\mathcal{G}$. We will use $\mathcal{D}$ to train a latent world model $\mathcal{W}$ (Sec.~\ref{sec:latent_world_model_learning}) and optimize the latent residual policy using $\mathcal{W}$ and $\mathcal{G}$ (Sec.~\ref{sec:residual_policy_training}). We also include a branch-start sampling strategy to improve the training efficiency in Sec.~\ref{sec:progress_conditioned_curriculum}.

\subsection{Learning Latent World Models with Hierarchical Forcing}
\label{sec:latent_world_model_learning}
\begin{figure*}[t]
    \centering
    \vspace{0em}
    \includegraphics[width=0.75\linewidth]{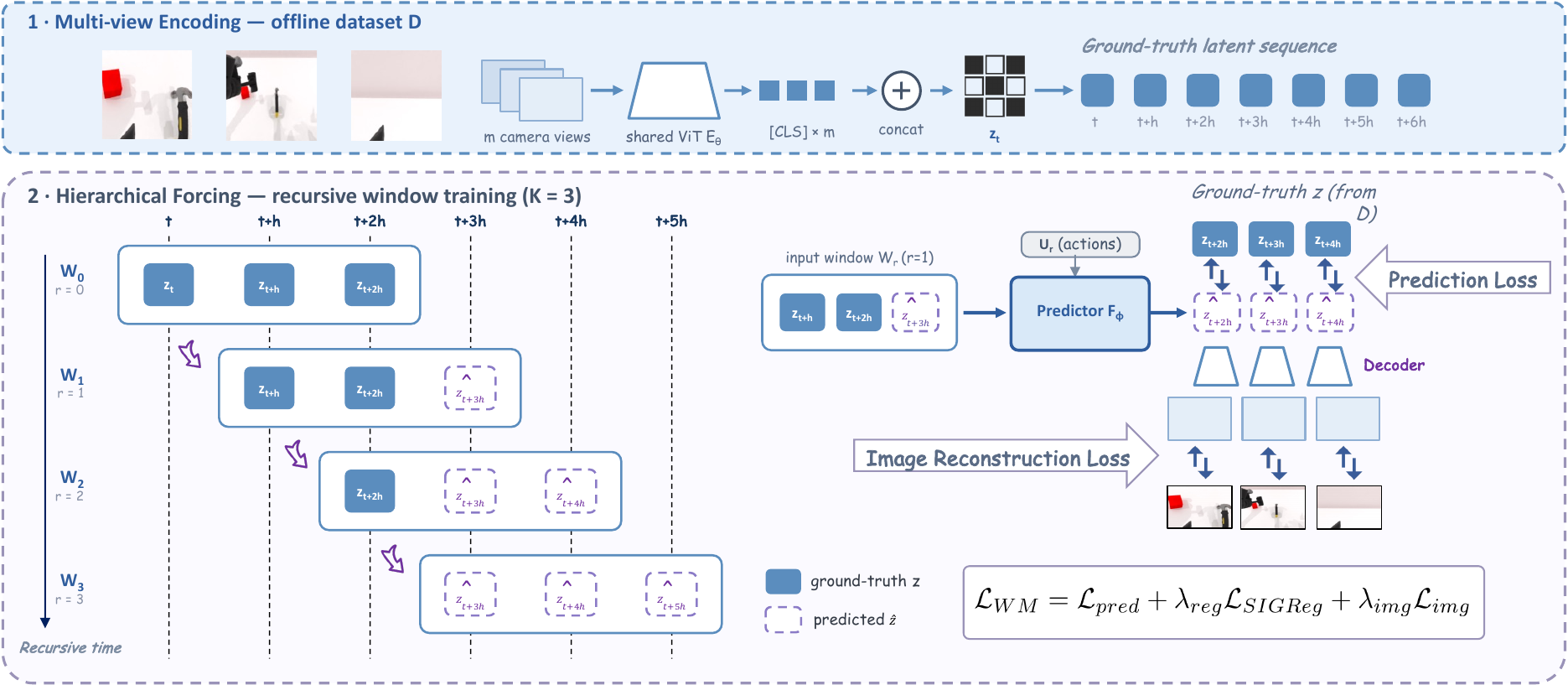}
    \vspace{0.5em}
    \caption{Training pipeline of the latent world model.}
    \label{fig:wm_training}
    \vspace{-1.5em}
\end{figure*}

We first describe how we learn a latent world model $\mathcal{W}$ from the offline dataset $\mathcal{D}$, which is shown in Fig.~\ref{fig:wm_training}. Following LeWorldModel, we employ a Vision Transformer (ViT) encoder to map visual observations into a latent representation space. Unlike conventional single-view settings, however, robotic observations typically consist of multiple camera views. In \methodname, given an observation containing $m$ views, denoted by $o_t = \{I_t^k\}_{k=1}^{m}$, we independently encode each view using the shared visual encoder and concatenate the features corresponding to the [CLS] tokens to form a joint latent representation:

\begin{equation*}
z_t =
\operatorname{Concat}_{k=1}^{m}
\left(
E_\theta(I_t^k)
\right)
\end{equation*}

Given the latent representations, we learn an action-conditioned dynamics model to predict future states directly in the latent space. We employ an action-conditioned predictor $F_\phi$ that takes a history of latent representations and the corresponding actions as input, and predicts the next-step latent representations.
Prior works \cite{zhou2025dinowm,LeWM} typically train the predictor using one-step prediction objectives. However, training a latent world model for imagination requires more than accurate one-step prediction. During imagined rollouts, the model must recursively predict future states by taking its own predictions as inputs. In contrast, conventional one-step training primarily exposes the predictor to ground-truth latent histories, creating a discrepancy between the training and inference conditions. As the model recursively feeds its predictions back as inputs, prediction errors may accumulate and degrade the reliability of imagined trajectories.

To bridge this gap, we introduce \textit{Hierarchical Forcing}, which combines dense causal prediction supervision within each temporal window with autoregressive feedback across consecutive windows.

Specifically, we initialize the prediction process with a window of $K$ consecutive ground-truth latent representations under the frame skip of $h$:
\begin{equation*}
W_0 = [z_t,z_{t+h},\cdots,z_{t+(K-1) \cdot h}]
\end{equation*}
At the $r$-th recursive step, the predictor takes the current latent window $W_r$ and its corresponding action conditions $U_r$ as input, and predicts the next-step latent representations at all positions:
\begin{equation*}
\hat{Y}_r = F_\phi(W_r,U_r)= [\hat{z}_{t+(r+1) \cdot h}^{\langle r\rangle},\cdots, \hat{z}_{t+(r+K) \cdot h}^{\langle r\rangle}]
\end{equation*}
As recursive prediction proceeds, the input window gradually incorporates model predictions. We denote the latent representation at each position by $\tilde{z}$, which can be either a ground-truth or a predicted latent:
\begin{equation*}
\tilde{z}_{t+(r+i) \cdot h} =
\begin{cases}
z_{t+(r+i) \cdot h}, & r+i < K\\
\hat{z}_{t+(r+i) \cdot h}^{\langle r+i-K \rangle}, & r+i \geq K
\end{cases}
\end{equation*}
Accordingly, the input window at the $r$-th recursive step is expressed as
\begin{equation*}
W_r =
[\tilde{z}_{t+r \cdot h},\cdots,
\tilde{z}_{t+(r+K-1) \cdot h}]
\end{equation*}
We employ causal attention in the predictor, ensuring that each position only attends to its current and preceding latent representations. This enables the predictor to perform next-step prediction at every position within the window.

We supervise the predictions at all positions within each window using their corresponding ground-truth latent representations from the offline trajectories. The prediction objective can be formulated as
\begin{equation*}
\mathcal{L}_{\mathrm{pred}}
=
\frac{1}{RK}
\sum_{r=0}^{R-1}
\sum_{i=0}^{K-1}
\left\|
\hat{Y}_r[i]-z_{t+(r+i+1) \cdot h}
\right\|_2^2
\end{equation*}
where $R$ denotes the number of recursive prediction steps. By supervising all positions within each window, our approach provides dense learning signals for latent dynamics prediction, rather than relying solely on the prediction at the final position.

As illustrated in Fig.~\ref{fig:wm_training}, the input window gradually transitions from entirely ground-truth representations to fully predicted representations as recursive prediction proceeds, while dense next-step supervision is maintained at every position. This allows the predictor to progressively adapt to the autoregressive input conditions encountered during imagined rollouts.

Finally, following LeWorldModel, we incorporate SIGReg to prevent representation collapse. We additionally train a CNN decoder to reconstruct images from every predicted latent in the output window at each recursive step. The reconstructed images may have limited visual fidelity, as illustrated in the qualitative results presented in Appx.~\ref{sec:example_rollouts}. We therefore use the decoder solely to provide auxiliary supervision that encourages the latent representations to retain visual information, rather than to generate observations for policy learning.
The overall training objective is
\begin{equation}
\label{eq:wm_loss}
\mathcal{L}_{\mathrm{WM}}
=
\mathcal{L}_{\mathrm{pred}}
+
\lambda_{\mathrm{reg}}\mathcal{L}_{\mathrm{SIGReg}}
+
\lambda_{\mathrm{img}}\mathcal{L}_{\mathrm{img}}
\end{equation}
where $\mathcal{L}_{\mathrm{img}}$ denotes the image reconstruction loss. The encoder, predictor, and decoder are jointly optimized, with gradients propagated through both the reconstruction and recursive feedback paths.

\subsection{Residual Policy Optimization via Latent Subchunk Rollouts}
\label{sec:residual_policy_training}
To optimize VLA policies using imagined trajectories generated by the latent world model, we need to enable closed-loop policy decisions within latent imagination. However, directly integrating pretrained VLAs with latent world models is non-trivial due to their inherent interface mismatch: while latent world models predict future states in a compact latent space, pretrained VLAs typically require visual observations as input to generate actions. Consequently, the predicted latent states cannot be directly fed into the original VLA to continue the rollout. To address this issue, we introduce a lightweight residual policy that directly takes the latent states predicted by the world model as input to do \textit{Latent Subchunk Rollouts}.

Our key idea is to decouple the provision of behavioral priors from state-dependent action refinement. To this end, we keep the pretrained VLA $\pi_{\mathrm{VLA}}$ frozen and use it to generate a base action chunk of length $H$ at the branch starting point:
\begin{equation*}
B_t = \pi_{\mathrm{VLA}}(o_t,\ell)
= [a_t^0,\ldots,a_{t+H-1}^0]
\end{equation*}
We then divide the base action chunk into $N$ shorter subchunks, each containing $h$ consecutive actions, such that $H=Nh$. The $j$-th base action subchunk is denoted as
\begin{equation*}
B_t^{(j)}
=
[a_{t+jh}^0,\ldots,a_{t+(j+1)h-1}^0],
\quad j=0,\ldots,N-1
\end{equation*}

At each subchunk decision step, the residual policy takes the current latent state, the corresponding base action subchunk, and the language instruction as input to generate an action correction:
\begin{equation*}
\Delta A_j
\sim
\pi_\psi
\left(
\cdot \mid
\hat{z}_{t+jh},B_t^{(j)},\ell
\right)
\end{equation*}
The corrected action subchunk is obtained by
\begin{equation}
\label{eq:action_refinement}
\widetilde{A}_j
=
\operatorname{clip}
\left(
B_t^{(j)}+\beta\Delta A_j, A_{min}, A_{max}
\right)
\end{equation}
where $\beta$ controls the magnitude of the residual correction and $\operatorname{clip}(\cdot)$ constrains the resulting actions to the valid action range. Here, we initialize $\hat{z}_t=z_t$, while subsequent latent states are predicted by the world model.

After obtaining the corrected action subchunk, we feed it into the latent world model to predict the subsequent latent state. Since the world model is trained with a frame skip of $h$, each prediction corresponds to the transition over one action subchunk:
\begin{equation*}
\hat{z}_{t+(j+1)h}
=
\left[
F_\phi
\left(
\widehat{\mathcal{H}}_j,
\widehat{A}_j
\right)
\right]_{K-1}
\end{equation*}
where $\widehat{\mathcal{H}}_j$ and $\widehat{A}_j$ denote the current latent history window and its corresponding action conditions, respectively. We take the prediction at the last position of the output window as the next latent state.
The predicted latent state is then used to condition the residual policy for refining the next base action subchunk. This process forms a closed-loop interaction between the residual policy and the latent world model, following an iterative procedure of action refinement and latent state prediction.

To provide task-level rewards for imagined trajectories, we train a lightweight success detector $\xi$ using datasets $\mathcal{G}$ and $\mathcal{D}$. We construct positive and negative training samples based on the temporal positions within each trajectory. We treat the last $n_1$ frames of trajectories in $\mathcal{G}$ as positive samples, while frames preceding the last $n_2$ frames of trajectories in both $\mathcal{G}$ and $\mathcal{D}$ are labeled as negative samples. The success detector is trained to distinguish these two types of states based on their latent representations and task descriptions: 
\begin{equation*}
\hat{q}_t = \xi(z_t,\ell)
\end{equation*}
During imagined rollouts, we evaluate the predicted latent state after each subchunk and assign a binary reward according to a predefined threshold $\tau$:
\begin{equation}
\label{eq:success_reward}
\hat{r}_j =
\mathbb{I}
\left[
\xi(\hat{z}_{t+(j+1)h},\ell) \geq \tau
\right]
\end{equation}
where $\hat{z}_{t+(j+1)h}$ denotes the predicted latent state after executing the $j$-th corrected action subchunk. The reward is set to $1$ if the predicted state satisfies the success criterion and $0$ otherwise.

We then optimize the residual policy using TD3+BC \cite{fujimoto2021minimal}, following the action refinement paradigm of RL Token \cite{rlt}. The imagined transitions are used to update the residual actor and critic, enabling policy optimization with additional training experience without requiring the VLA to be updated or repeatedly queried during imagined rollouts.

\subsection{Progress-Conditioned Branch-Start Curriculum}
\label{sec:progress_conditioned_curriculum}
Although latent subchunk rollouts provide an efficient way to generate imagined experiences, their limited rollout horizon may prevent branches starting from early task stages from reaching successful states. This makes it difficult for the residual policy to obtain informative success feedback, particularly during early training. To address this issue, we introduce a \textit{\textbf{P}rogress-conditioned \textbf{B}ranch-start \textbf{C}urriculum} (PBC), which gradually shifts the sampling distribution of imagined rollout starting points from later task stages to earlier ones.

We train a progress predictor $g_p$ using the expert dataset $\mathcal{G}$ to estimate task completion levels from latent representations and corresponding language instructions:
\begin{equation*}
\hat{p}_t = g_p(z_t,\ell)
\end{equation*}
where $\hat{p}_t$ denotes the estimated progress at time step $t$. We use the predicted progress exclusively for branch-start sampling rather than directly as a reward signal. The progress predictor is trained on expert demonstrations, making its predictions unreliable on states induced by imperfect or exploratory policies. Using such predictions as dense rewards may introduce exploitable estimation errors, allowing the policy to optimize the predictor's outputs rather than make genuine task progress. In contrast, branch-start sampling does not require highly accurate progress estimates, as it primarily relies on coarse task-stage information. For example, a state with an actual progress of $0.9$ may be predicted as $0.7$, but it can still be recognized as a relatively late-stage state and assigned a higher sampling probability.

To be specific, let $\mathcal{K}_{\mathrm{start}}$ denote the set of candidate branch starting states sampled from trajectories collected through environment interaction. At training epoch $e$, we construct a progress-conditioned sampling distribution:
\begin{equation}
\label{eq:pbc_sampling}
P_e(i)
=
\frac{w_e(\hat{p}_i)}
{\sum_{j\in\mathcal{K}_{\mathrm{start}}}w_e(\hat{p}_j)}
\end{equation}
where $w_e(\cdot)$ denotes a progress-dependent weighting function that evolves throughout training. In the early epochs, higher-progress states receive larger sampling weights, allowing short imagined rollouts to more frequently reach task completion and obtain successful feedback. As training progresses, the sampling distribution gradually expands toward lower-progress states, encouraging the residual policy to improve its behavior over increasingly earlier stages of the task. 
\section{Experiments}
In this section, we conduct extensive experiments to evaluate our proposed method~\methodname. We first introduce the experiment setup.

\subsection{Experiment Setup}
\paragraph{Benchmarks.} We conduct our experiments on two robotic manipulation benchmarks: LIBERO~\cite{libero} and RoboTwin~\cite{robotwin}. 
(1) \textbf{LIBERO}: LIBERO is a widely adopted benchmark consisting of four task suites: \textit{Spatial}, \textit{Object}, \textit{Goal}, and \textit{Long}. The offline dataset $\mathcal{D}$ for LIBERO contains 200 trajectories for each task, which are collected by a $\pi_{0.5}$ SFT checkpoint with random noise added to the generated action chunks. The expert dataset $\mathcal{G}$ contains 10 episodes for each task. We evaluate 50 episodes for each task in four task suites.
(2) \textbf{RoboTwin}: RoboTwin is a large-scale benchmark for bimanual robotic manipulation. In our experiments, we select three representative tasks for evaluation: \textit{Adjust Bottle}, \textit{Beat Block Hammer} and \textit{Click Alarmclock}. The offline dataset $\mathcal{D}$ for RoboTwin contains 800 trajectories for each task collected in a similar way to LIBERO. The expert dataset $\mathcal{G}$ contains 50 episodes for each task.
We evaluate 80 episodes for each RoboTwin task.
(3) \textbf{ALOHA}: We include a real-world ALOHA robot to do two tasks: \textit{Pen into Holder} and \textit{Plug Charger}. The offline dataset $\mathcal{D}$ contains 400 trajectories for each task while the expert dataset $\mathcal{G}$ consists of 100 episodes for each task. We evaluate 20 episodes for each ALOHA task.

\paragraph{Baselines.} Since our method adopts a residual policy attached to a frozen pretrained VLA for RL optimization, we follow the same design principle for all other baselines to ensure a fair comparison. Comparisons with methods that update the full VLA parameters are deferred to the ablation studies.
The baselines are as follows:
(1) \textbf{$\pi_{0.5}$-SFT}: The base model trained using SFT for~\methodname~along with the other baselines.
(2) \textbf{RLT-BC}: A residual policy which takes RL token as the task feature input and is trained with BC. Unlike the original RLT, we follow the approach adopted in VLARLKit \cite{vlarlkit2026} and directly apply mean pooling to the PaliGemma embeddings to obtain the RL Token.
(3) \textbf{Latent-BC}:A residual policy which takes the world model latent of $\mathcal{W}$ as the task feature input and is trained with BC. 
(4) \textbf{RLT-RL}: A residual policy which takes RL token as the task feature input and is trained using online RL.
(5) \textbf{Latent-RL}: A residual policy which takes the world model latent of $\mathcal{W}$ as the task feature input and is trained using online RL.
(6) \textbf{Dreamer}: A latent MBRL baseline that trains the latent world model in a Dreamer~\cite{DreamerV3}~style and uses subchunk rollouts for residual policy optimization.

\paragraph{Implementation Details.} For the residual policies trained by both \methodname~and the baselines, we adopt the following inference procedure: the VLA generates a base action chunk of length \(H\), and the residual policy refines its first \(h\) actions. These refined actions are executed in the environment before querying the VLA again with the updated observation. 
During imagined rollouts in \methodname, we instead reuse a single VLA-generated base action chunk and successively refine its subchunks as described in Sec.~\ref{sec:residual_policy_training}. We denote the actual imagined rollout horizon by $L \leq H$, measured in environment steps, where $L$ is a multiple of $h$. Each branch therefore processes the first $L/h$ subchunks without re-querying the VLA. We use \(h=5\), \(H=50\), and \(L=50\) for all tasks unless otherwise specified.

During the training of MBRL algorithms, including \methodname~and Dreamer, the residual policy interacts with the simulator to collect 10 trajectories per epoch, following the rollout procedure described above. These 10 trajectories are then used as starting points for branch rollouts within the corresponding epoch. After completing the branch rollouts, we perform RL updates using the collected rollout data. The algorithms are trained for a total of 10 epochs.
For RLT-RL and Latent-RL, we ensure the same number of simulator interactions as \methodname~and Dreamer, resulting in a total of 100 trajectories.

\begin{table*}[t]
\centering
\caption{Performance comparison on LIBERO and RoboTwin benchmarks.}

\label{tab:main_results}
% \footnotesize
% \setlength{\tabcolsep}{4pt}
% \renewcommand{\arraystretch}{1.15}
\small
\setlength{\tabcolsep}{5pt}
\renewcommand{\arraystretch}{1.2}

\begin{tabular}{
lcccc
!{\vrule width 0.8pt}
>{\centering\arraybackslash}p{1.15cm}
>{\centering\arraybackslash}p{1.25cm}
>{\centering\arraybackslash}p{1.25cm}
}
\toprule
\multirow{2}{*}{\textbf{Model}}
& \multicolumn{4}{c}{\textbf{LIBERO}}
& \multicolumn{3}{c}{\textbf{RoboTwin}} \\
\cmidrule(lr){2-5}
\cmidrule(lr){6-8}
& \textbf{Spatial}
& \textbf{Object}
& \textbf{Goal}
& \textbf{Long}
& \shortstack{\textbf{Adjust}\\\textbf{Bottle}}
& \shortstack{\textbf{Beat Block}\\\textbf{Hammer}}
& \shortstack{\textbf{Click}\\\textbf{Alarmclock}} \\
\midrule
$\pi_{0.5}$-SFT & 78.4 & 87.2 & 73.6 & 52.0 & 90.0 & 82.5 & 51.3 \\
RLT-BC          & 83.2 & 88.8 & 74.4 & 59.6 & 92.5 & 80.0 & 63.8 \\
Latent-BC       & 84.0 & 83.2 & 75.2 & 68.0 & 91.3 & 85.0 & 67.5 \\
RLT-RL          & 86.4 & 92.0 & 76.8 & 68.8 & 93.8 & 87.5 & 72.5 \\
Latent-RL       & 85.6 & 90.4 & 75.2 & 66.2 & 91.3 & 86.3 & 68.8 \\
Dreamer         & 88.0 & 92.8 & 77.6 & 69.6 & 95.0 & 92.5 & 80.0 \\
\midrule
\textbf{\methodname}
& \textbf{92.0} & \textbf{96.0} & \textbf{84.8} & \textbf{71.2}
& \textbf{97.5} & \textbf{95.0} & \textbf{83.8} \\
\bottomrule
\end{tabular}
\end{table*}

\subsection{Main Results}
\paragraph{Simulation Results.} Tab.~\ref{tab:main_results} shows the average success rate of different approaches on LIBERO and RoboTwin benchmarks. 
In general, our proposed method \methodname~outperforms the baselines for all task environments. 
Since all baselines use $\pi_{0.5}$-SFT as the base VLA, most of them achieve improved success rates compared with the base model. For RLT-BC and Latent-BC, we observe that Latent-BC generally performs better than RLT-BC across most tasks. We attribute this phenomenon to the potentially more task-relevant representations provided by our world model compared with the RL Token representations, which may facilitate better generalization during behavior cloning. However, when switching from behavior cloning to online RL, the performance gap between the two approaches becomes less pronounced. We conjecture that this is because online RL allows policies to explore and learn through trial and error, potentially reducing their reliance on the quality of the underlying representations.
Both MBRL approaches, namely \methodname~and Dreamer, outperform the BC-based and online RL baselines in most tasks. This improvement can be attributed to the additional imagined trajectories generated by the latent world model, which provide extra training experiences beyond real-env interactions under the same interaction budget.
Furthermore, \methodname~achieves better performance than Dreamer. We think this might be caused by the combination of Hierarchical Forcing and SIGReg, which facilitates learning more effective latent representations and mitigates compounding errors during recursive world model rollouts, thereby improving the quality of imagined experiences.

\paragraph{Real-world Results.}
We further evaluate our method on real-world manipulation tasks using an ALOHA robot. As shown in Fig.~\ref{fig:aloha_task}, both Dreamer and \methodname~achieve higher success rates than the SFT-trained $\pi_{0.5}$ baseline through MBRL with latent subchunk rollouts. Moreover, the Hierarchical Forcing helps mitigate the compounding error of the latent world model, which empowers \methodname~to outperform Dreamer in both tasks.

\subsection{Effectiveness of Hierarchical Forcing}
% \paragraph{Effect of Hierarchical Forcing.}
To evaluate the effectiveness of Hierarchical Forcing, we compare the latent world model in \methodname~with:
(1) \textbf{LeWM}, a LeWM-style baseline~\cite{LeWM} trained with one-step latent prediction under teacher forcing and auxiliary multi-view image reconstruction; and
(2) \textbf{Recursive-only}, which uses the same recursive feedback as Hierarchical Forcing but applies the latent prediction loss only to the final output position of each window.
All variants use the same model architecture and auxiliary loss terms. 
We evaluate the three world models on 10 held-out trajectories from LIBERO Long, using identical initial image observations and recorded action sequences. Each model recursively predicts 100 environment steps, corresponding to 20 subchunks.
We report two metrics:
(1) \textbf{PSNR-FINAL}, evaluated at the final prediction step; and
(2) \textbf{PSNR-ALL}, averaged over all prediction steps.
Both metrics compare decoded images with their corresponding ground-truth observations.
The results are reported in Tab.~\ref{tab:ablation_hierarchical_forcing}.
Hierarchical Forcing achieves the highest PSNR-FINAL and PSNR-ALL, indicating improved decoded-image fidelity during recursive rollouts.
More qualitative comparisons are provided in Appx.~\ref{sec:example_rollouts}.

\begin{table}[t]
\centering
\caption{Ablation study of Hierarchical Forcing on LIBERO Long.
PSNR is reported in dB, and Policy denotes the success rate (\%).}
\label{tab:ablation_hierarchical_forcing}
\small
\setlength{\tabcolsep}{8pt}
\renewcommand{\arraystretch}{1.15}

\begin{tabular}{
lcc
!{\vrule width 0.8pt}
c
}
\toprule
\textbf{Method}
& \textbf{PSNR-FINAL}
& \textbf{PSNR-ALL}
& \textbf{Policy} \\
\midrule
LeWM
& 15.53 & 18.51
& 61.4 \\

Recursive-only
& 23.84 & 25.65
& 68.0 \\

\textbf{\methodname}
& \textbf{25.22} & \textbf{27.90}
& \textbf{71.2} \\
\bottomrule
\end{tabular}
\vspace{-1.5em}
\end{table}

Furthermore, we train separate residual policies using subchunk rollouts generated by the three world models, while keeping the base VLA and policy optimization protocol identical.
The corresponding success rates are reported in the \textbf{Policy} column of Tab.~\ref{tab:ablation_hierarchical_forcing}.
Compared with Recursive-only, Hierarchical Forcing improves policy success from 68.0\% to 71.2\%.
Together, these results support the benefit of dense within-window latent prediction supervision for both recursive prediction quality and downstream residual policy learning.

\subsection{Effectiveness of Latent Subchunk Rollouts}
\begin{figure}[t]
    \centering
    \subfigure[Real-world results on ALOHA tasks.]{%
        \includegraphics[width=0.48\linewidth]{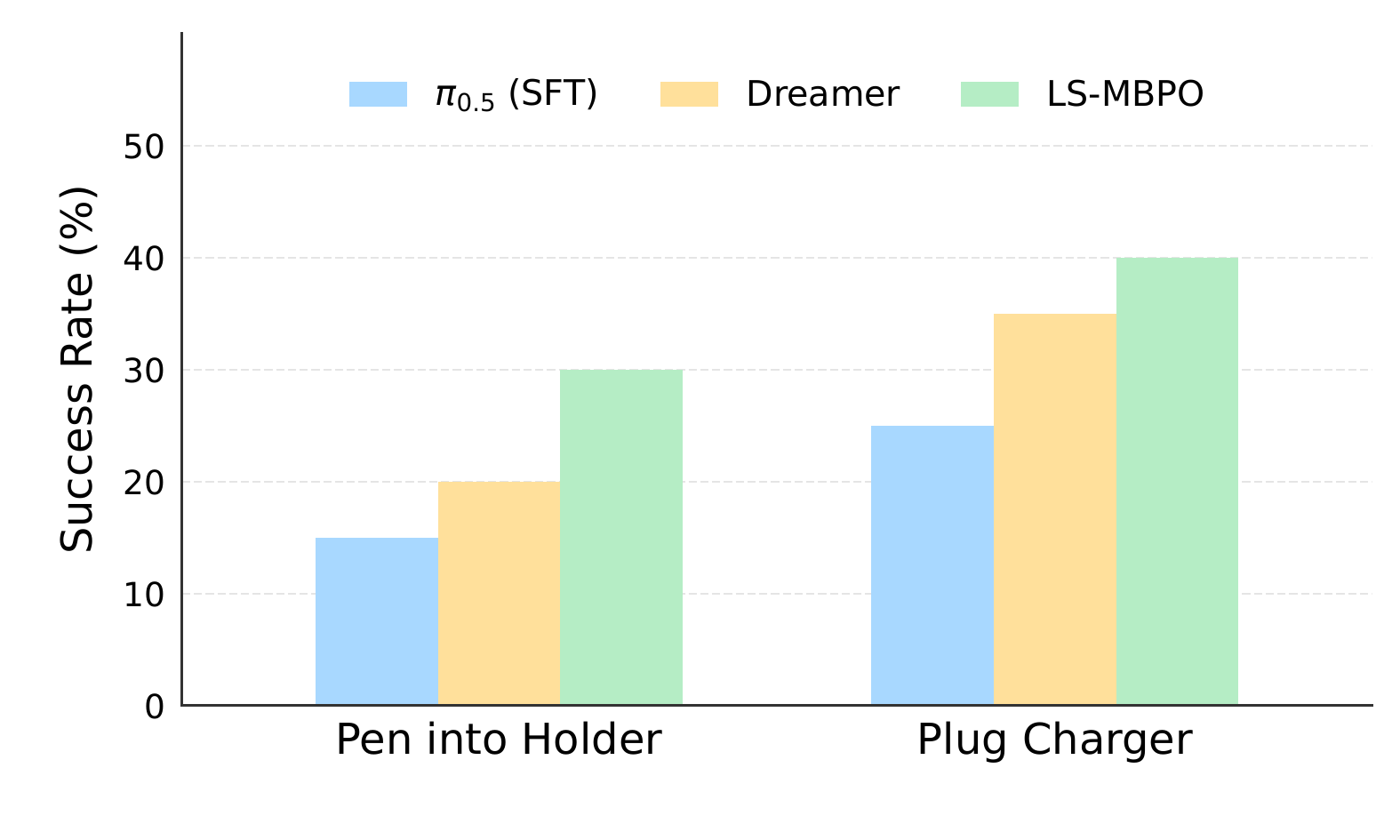}%
        \label{fig:aloha_task}%
    }\hfill%
    \subfigure[Ablation results of \methodname~on LIBERO Long tasks.]{%
        \raisebox{4mm}[\height][\depth]{%
            \includegraphics[width=0.48\linewidth]{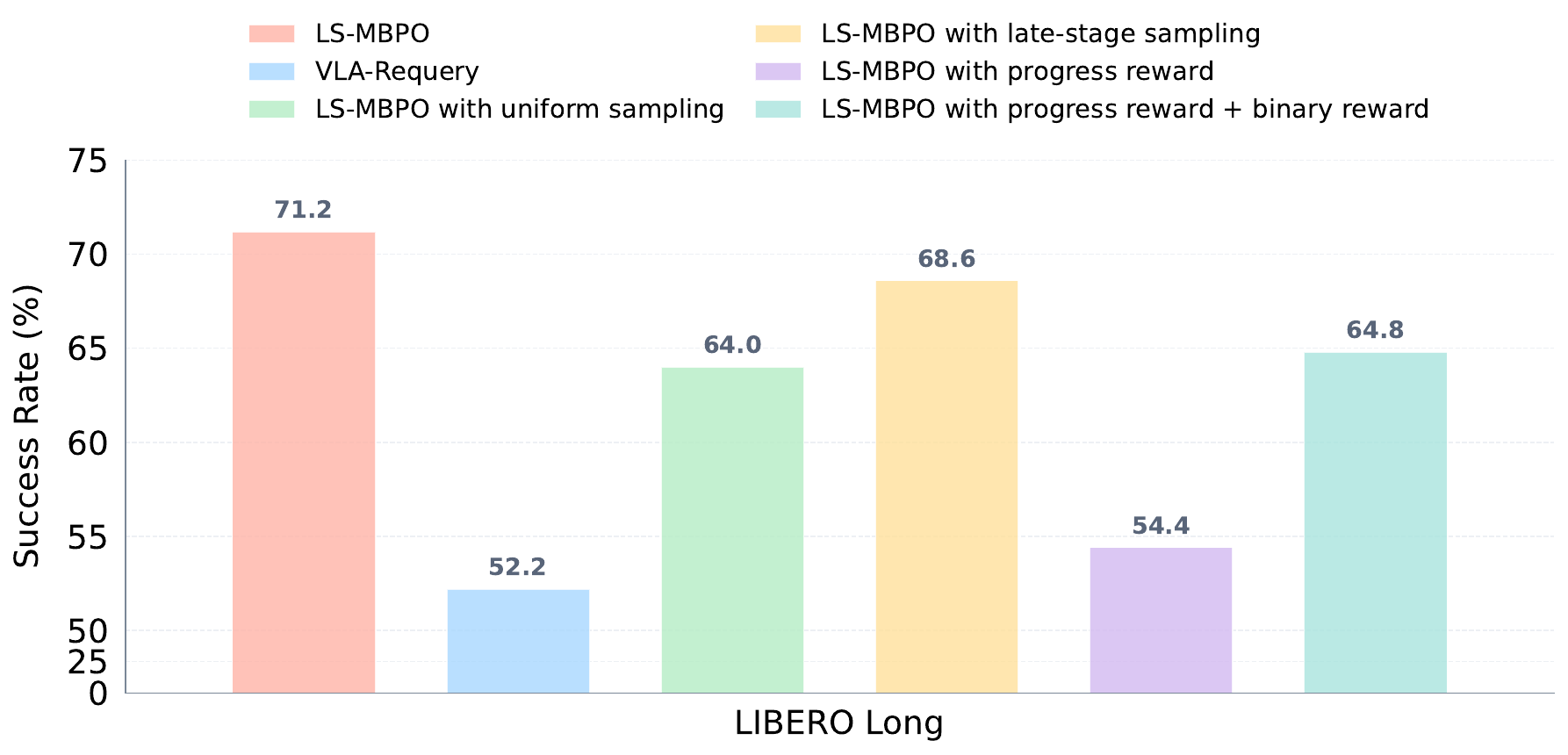}%
        }%
        \label{fig:libero_long_ablation}
    }

    \caption{Real-world evaluation on ALOHA and ablation study on LIBERO Long.}
    \label{fig:aloha_libero_combined}
\end{figure}

Since our latent world model is trained with an auxiliary decoder, it is also possible to perform recursive rollouts by re-querying the VLA with reconstructed observations. After each corrected $h$-step subchunk, we decode the predicted latent state into multi-view observations and query the frozen VLA to generate a new base action chunk. The residual policy then refines the first $h$ actions of this chunk to continue the imagined rollout.
This process can be repeated to enable closed-loop recursive rollouts. 
The performance is reported in Fig.~\ref{fig:libero_long_ablation}, where \methodname~outperforms this decoder-based VLA requery approach, which we call \textit{VLA-Requery} in the figure. Two factors, in our perspective, contribute to the performance drop. First, in our latent subchunk rollouts, the VLA already provides a complete base action chunk that serves as a strong behavioral prior for multiple successive residual decisions, making repeated VLA inference unnecessary. Second, image reconstruction inevitably introduces extra errors, which may further affect the actions generated by the VLA when conditioned on reconstructed observations.

\subsection{Ablation Study}
\label{sec:ablation}

\paragraph{Effect of PBC.}
To evaluate the effectiveness of Progress-Conditioned Branch-Start Curriculum (PBC), we compare it with two alternatives: (1) \textbf{Uniform sampling}, which samples branch starting states uniformly; and (2) \textbf{Late-stage sampling}, which favors later-stage states using a fixed sampling rule throughout training. All variants are trained for the same number of epochs. As shown in Fig.~\ref{fig:libero_long_ablation}, both uniform sampling and late-stage sampling achieve lower success rates than PBC. These results suggest that prioritizing later-stage states benefits policy learning, while gradually expanding the branch-start distribution toward earlier task stages yields further improvement.

\paragraph{Binary Success Reward Versus Progress Reward.} 
Our method uses binary success rewards (0/1) to optimize the residual policy, while the progress predictor is exclusively used for PBC. A natural question is whether replacing the binary reward with progress-based rewards could provide denser learning signals and potentially lead to better performance. To investigate this, we conduct two additional comparison experiments: (1) using the progress predictor as the reward function; (2) using the combination of progress reward and binary reward. However, the results, seen in Fig.~\ref{fig:libero_long_ablation}, show that progress-based rewards do not yield satisfactory performance. We consider this is caused by limited generalization capability of the progress predictor, which is trained on a small set of expert demonstrations and may produce unreliable estimates on states outside the expert distribution. As the policy explores such states, inaccurate progress predictions can introduce misleading reward signals and distort policy updates.

\begin{table}[t]
\centering
\caption{Comparison with VLA-MBPO on LIBERO Long. LS-MBPO achieves higher success rate while using fewer computational resources and training time.}
\vspace{0.5em}
\label{tab:ablation_vla_mbpo}
\footnotesize
\setlength{\tabcolsep}{2.5pt}
\renewcommand{\arraystretch}{1.15}

\begin{adjustbox}{max width=\columnwidth}
\begin{tabular}{@{}lccc!{\vrule width 0.8pt}cc@{}}
\toprule
\multirow{2}{*}{\textbf{Method}}
& \multirow{2}{*}{\shortstack{\textbf{Success}\\\textbf{Rate (\%)}}}
& \multicolumn{2}{c!{\vrule width 0.8pt}}{\textbf{Training Time}}
& \multicolumn{2}{c@{}}{\textbf{GPU Configuration}} \\
\cmidrule(lr){3-4}
\cmidrule(lr){5-6}
&
& \textbf{WM}
& \textbf{RL}
& \textbf{WM}
& \textbf{RL} \\
\midrule
VLA-MBPO
& 64.4
& 18\,h
& 12\,h
& \shortstack{4 $\times$ H100\\(80\,GB each)}
& \shortstack{4 $\times$ H100\\(80\,GB each)} \\

\midrule

\textbf{\methodname}
& \textbf{71.2}
& \textbf{12\,h}
& \textbf{1.5\,h}
& \shortstack{\textbf{4 $\times$ RTX 5090}\\(32\,GB each)}
& \shortstack{\textbf{1 $\times$ RTX 5090}\\(32\,GB each)} \\
\bottomrule
\end{tabular}
\end{adjustbox}
\vspace{-1.5em}
\end{table}

\paragraph{Comparison to VLA-MBPO.} 
Our method enables lightweight MBRL for VLA policies by combining latent world models with residual policy learning. We compare \methodname~against VLA-MBPO, which employs a large-scale generative world model and fine-tunes the full VLA parameters, on LIBERO Long. The results are reported in Tab.~\ref{tab:ablation_vla_mbpo}. Notably, \methodname~achieves even better performance than VLA-MBPO with much less computational resource. We think this improvement is because of the task-relevant representations learned by our latent world model, which transform policy optimization from the high-dimensional image space into a more compact latent space, potentially simplifying the learning problem and facilitating more effective policy optimization.

\section{Conclusion}
This study explores how to apply a latent world model to improve VLA through RL optimization. We propose a novel approach, \methodname, which optimizes a residual policy attached to pretrained VLA using imagined Subchunk Rollouts from a latent world model learned by Hierarchical Forcing. 
Extensive experiments demonstrate that \methodname~is capable of being applied to a variety of robotic platforms, enabling a robust and practical pathway towards lightweight model-based RL for robotics. 
Despite the promising results, \methodname~still leaves room for further exploration. 
First, future work can extend latent world models to support full-horizon rollouts for longer imagined interactions. 
Another potential direction is developing more generalizable dense reward models in latent space to provide richer feedback for policy optimization beyond binary success signals. Last but not least, future research can focus on learning generalizable latent world models from diverse tasks and robotic platforms to enable knowledge transfer and facilitate more scalable model-based policy learning across embodiments.
We believe these directions hold great potential for advancing latent world models and enabling more effective VLA training in the future.

% \clearpage

\bibliographystyle{apalike}
\bibliography{references}

\clearpage
\appendix

\renewcommand{\thepart}{}
\renewcommand{\partname}{}

\part{\huge{\textbf{Appendix}}} %
% \parttoc %

\label{sec:appendix}

\section{More Implementation Details}
\subsection{More Details for World Model Training}
\paragraph{The structure of decoder.}
For Image reconstruction in the latent world model, we reconstruct multi-view observations by splitting the predicted latent vector into view-specific components and decoding each component with an independently parameterized convolutional decoder.
An MLP followed by a linear projection maps each component to a $192 \times 16 \times 16$ feature map.
Four residual upsampling blocks progressively increase the spatial resolution to $256 \times 256$, with output channel widths of $96$, $48$, $32$, and $32$.
Each block performs $2\times$ nearest-neighbor upsampling followed by two $3\times3$ convolutions with Group Normalization and SiLU activations, while a final $1\times1$ convolution produces the three-channel image.

\paragraph{Proprioceptive prediction.}
In addition to image reconstruction, we train an auxiliary
proprioceptive decoder to predict the robot's proprioceptive
state from the full predicted latent representation.
At each recursive training step, this decoder is applied to
all output positions and supervised using the corresponding
ground-truth proprioceptive states.
During imagined rollouts, the next proprioceptive state is
decoded from the predicted next latent and used to update
the proprioceptive history provided to the residual actor
and critics.
For simplicity and clarity, proprioceptive inputs and their prediction are not included in the notation and figures in the main text of the paper.

\subsection{More Details for Policy Training}
\paragraph{Policy optimization details.}
Let $x_j$ denote the concatenation of the current latent state, proprioception, and language embedding at
subchunk step $j$.
The two critics take $(x_j,\widetilde A_j)$ as input and
estimate the expected task return of executing the corrected
action subchunk.
We store imagined transitions
$(x_j,B_t^{(j)},\widetilde A_j,\hat r_j,x_{j+1},d_j)$
in the replay buffer, retaining the reference subchunks
for actor conditioning. 

The critics are trained by regression to the TD target
\begin{equation}
y_j
=
\hat r_j
+
\gamma(1-d_j)
\min_{i=1,2}
Q_{\bar\theta_i}
\left(
x_{j+1},
\widetilde A_{j+1}^{\mathrm{targ}}
\right)
\end{equation}
where $\gamma$ is the discount factor per subchunk
transition and barred parameters denote target networks.
For nonterminal transitions,
$\widetilde A_{j+1}^{\mathrm{targ}}$ is constructed by
the target actor using $x_{j+1}$ and the next reference
subchunk $B_t^{(j+1)}$.
The indicator $d_j$ is set to one upon task success
or at the imagination horizon and otherwise zero.

\paragraph{Imagination boundary and value propagation.}
For intermediate subchunks ($j<N-1$), the next reference
subchunk $B_t^{(j+1)}$ is available within the VLA-generated
chunk. Together with $x_{j+1}$, it conditions the target
actor to produce the corrected action used for bootstrapping.
At the final subchunk, the reference chunk is exhausted,
so constructing the next target action would require a
reference continuation beyond the current imagination horizon.
We therefore set $d_{N-1}=1$, multiplying the continuation
term by zero and yielding
\begin{equation}
\label{eq:continuation_value}
y_{N-1}=\hat r_{N-1}
\end{equation}
This transition is trained against its immediate reward.
As branch-start states vary along real trajectories,
states near one imagined window's boundary can appear
inside another and receive nonterminal TD supervision.
These updates share the same critic, allowing success-related
value information to propagate to earlier task stages
through successive TD updates under PBC.

\paragraph{Behavioral regularization.}
The BC term anchors the actor to the corresponding
frozen-VLA reference subchunk:
\begin{equation}
\label{eq:bc_regularization}
\mathcal L_{\mathrm{BC}}
=
\mathbb E\left[
\frac{
\operatorname{MSE}
\left(
\widetilde A_\psi(x_j,B_t^{(j)}),
B_t^{(j)}
\right)
}{\beta^2}
\right]
\end{equation}
where $\widetilde A_\psi$ denotes the corrected action subchunk
produced by the current actor using
Eq.~\ref{eq:action_refinement}.
This term constrains deviations from the reference actions
to preserve the pretrained VLA's behavioral prior,
while the critic guides task-improving corrections.

\paragraph{RLT-BC and Latent-BC.}
We train the residual policies by minimizing the mean squared
error between corrected action subchunks defined in
Eq.~\ref{eq:action_refinement} and the corresponding expert
action subchunks sampled from the expert demonstration
dataset, keeping the VLA and feature extractors frozen.

\paragraph{RLT-RL and Latent-RL.}
We perform TD3-style actor--critic updates on replayed
transitions collected through online environment interaction.
The actor objective incorporates the BC regularizer
defined in Eq.~\ref{eq:bc_regularization}.
For Latent-RL, $\mathcal{W}$ remains frozen and serves
as the observation encoder.

\paragraph{Dreamer.}
We pretrain a DreamerV3 RSSM with reconstruction, KL,
reward, and continuation losses and freeze it during
policy optimization.
Each 50-step VLA-generated base action chunk is divided
into ten 5-step subchunks, with each corrected action
subchunk driving one RSSM transition.
The actor and state-value critic are trained using
DreamerV3's $\lambda$-return actor--critic objective
with entropy regularization.
The critic estimates state values from the RSSM state
and language embedding, allowing the final imagined state
to be bootstrapped without a subsequent
corrected action subchunk.
We therefore retain continuation-weighted value bootstrapping
at the imagination horizon like Eq.~\ref{eq:continuation_value} does.

\subsection{More Details about PBC}
Starting imagined branches from later task stages can reduce
the exploration required to encounter successful outcomes.
This allows the residual policy to first learn the local
corrective behaviors required for task completion. Meanwhile,
successful trajectories provide positive reward samples, i.e., the binary reward, for
the critic, helping it distinguish state-action pairs with
different expected task returns. This is particularly useful
during early training, when insufficient successful experience
may limit the critic's ability to establish meaningful value
differences.

As training proceeds, PBC gradually shifts the preferred
branch-start states toward earlier task stages. The policy
must then learn to reach and complete the later-stage behaviors
practiced previously. The critic can use its existing value
estimates for these later states to propagate success-related
information to earlier states through temporal-difference
updates. This curriculum is therefore intended to establish
task-completion behaviors and useful value estimates before
extending learning to earlier portions of the task.

\subsection{More Details about the Progress Predictor}
The progress reward evaluated in our ablation in Sec.~\ref{sec:ablation} is the
difference between consecutive progress predictions from the learned progress predictor. 
\begin{equation*}
r_t^{\mathrm{prog}}
=
\hat{p}_{t+1} - \hat{p}_t
\end{equation*}
This form of reward draws on the idea of reward shaping by treating the progress as a potential function and is
used either alone or in combination with the binary success
reward.

\subsection{More Details for Real-world Experiments}
\begin{figure*}[t]
    \centering
    \vspace{0em}
    \includegraphics[width=0.8\linewidth]{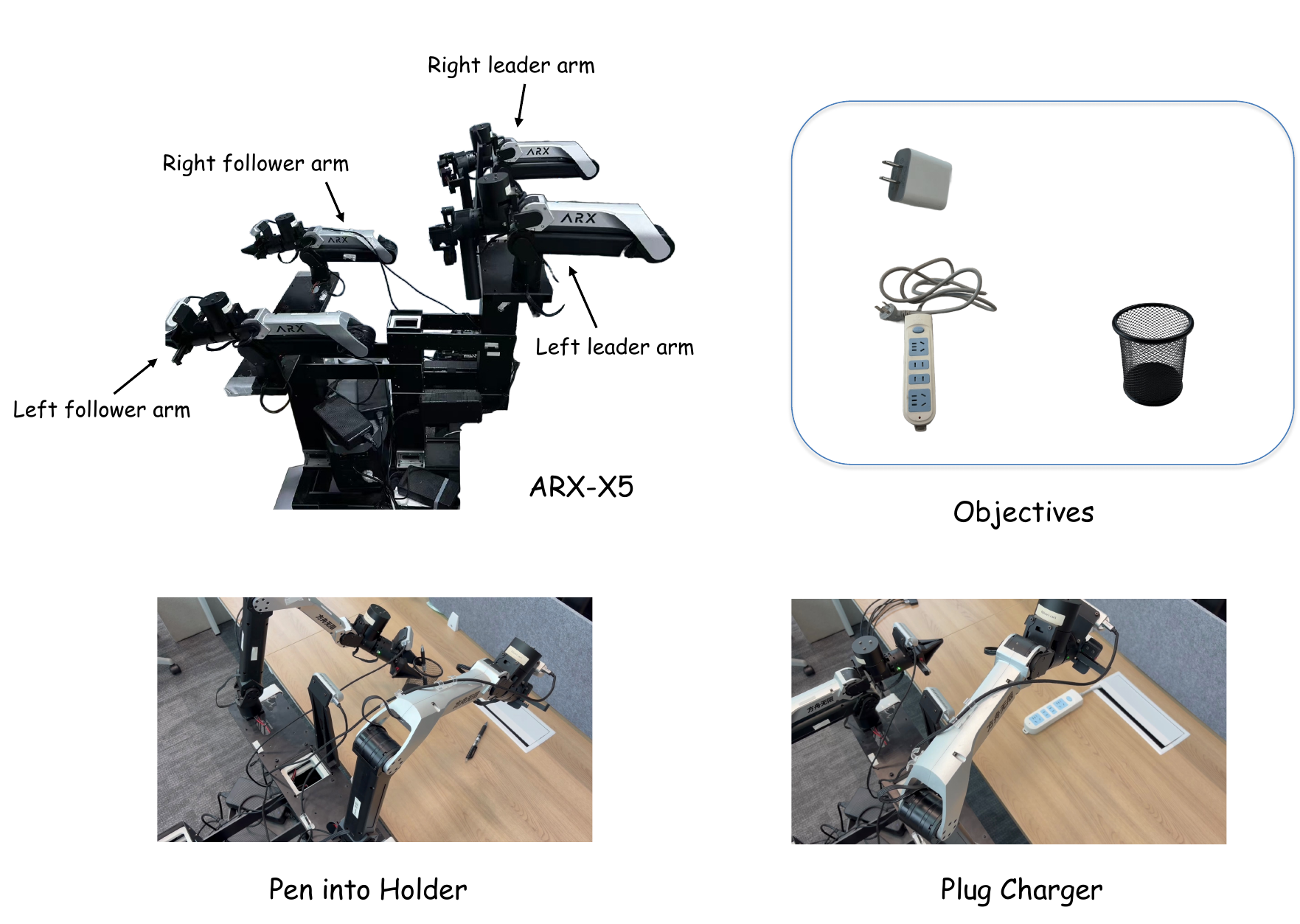}
    \vspace{-0.5em}
    \caption{The robot, objectives and the tasks used in real-world experiments.}
    \label{fig:real_world}
    \vspace{-1.5em}
\end{figure*}

Fig.~\ref{fig:real_world} illustrates our real-world experimental platform and task suite. All experiments are conducted on an ARX-X5 bimanual system consisting of two leader arms and two follower arms: during data collection, a human operator holds the leader arms and teleoperates the follower arms to provide expert demonstrations, whereas during evaluation the follower arms execute the policy fully autonomously. The evaluation involves two manipulation tasks built on daily objects—pens, a pen holder, a plug, and a power strip.

Specifically, Pen into Holder is a bimanual coordination task: the left arm first grasps and lifts the pen holder, the right arm then inserts the pens into it one by one, and the robot finally places the filled holder back onto the table. Plug Charger, in contrast, demands sub-centimeter precision to align and insert the plug into a socket of the power strip, probing the policy's fine-grained visual-motor control under high-precision alignment.

\section{More Experimental Results}
\subsection{Effect of VLA Prior}
In \methodname, the VLA provides base actions as behavioral priors, upon which the residual policy is optimized. To investigate the contribution of the VLA prior, we additionally train a latent policy without access to VLA-generated actions. Specifically, we keep the latent world model fixed and first pretrain the latent policy using behavior cloning (BC) on expert demonstrations. We then optimize this policy through branch-rollout MBRL within the latent world model. We consider two variants: one optimizes the policy using only the RL objective during MBRL, while the other additionally incorporates the BC loss as a regularization term.

The results are reported in Fig.~\ref{fig:appendix_libero_long_ablation}. We observe that, without BC regularization, the policy achieves very low success rates, likely because the challenging tasks make policy learning highly sensitive to deviations from the expert behavior. Once the policy deviates from successful behaviors, it may fail to discover successful trajectories through exploration, leading to further performance degradation. Adding BC regularization does yield improvements, but the performance is still below ours. In contrast, \methodname~achieves much better performance by leveraging the VLA to provide a stable behavioral prior, which guides residual policy learning and makes policy optimization more effective.

\subsection{Effect of Decoder}
\begin{figure}[t]
    \centering
    \vspace{0em}
    \includegraphics[width=0.5\linewidth]{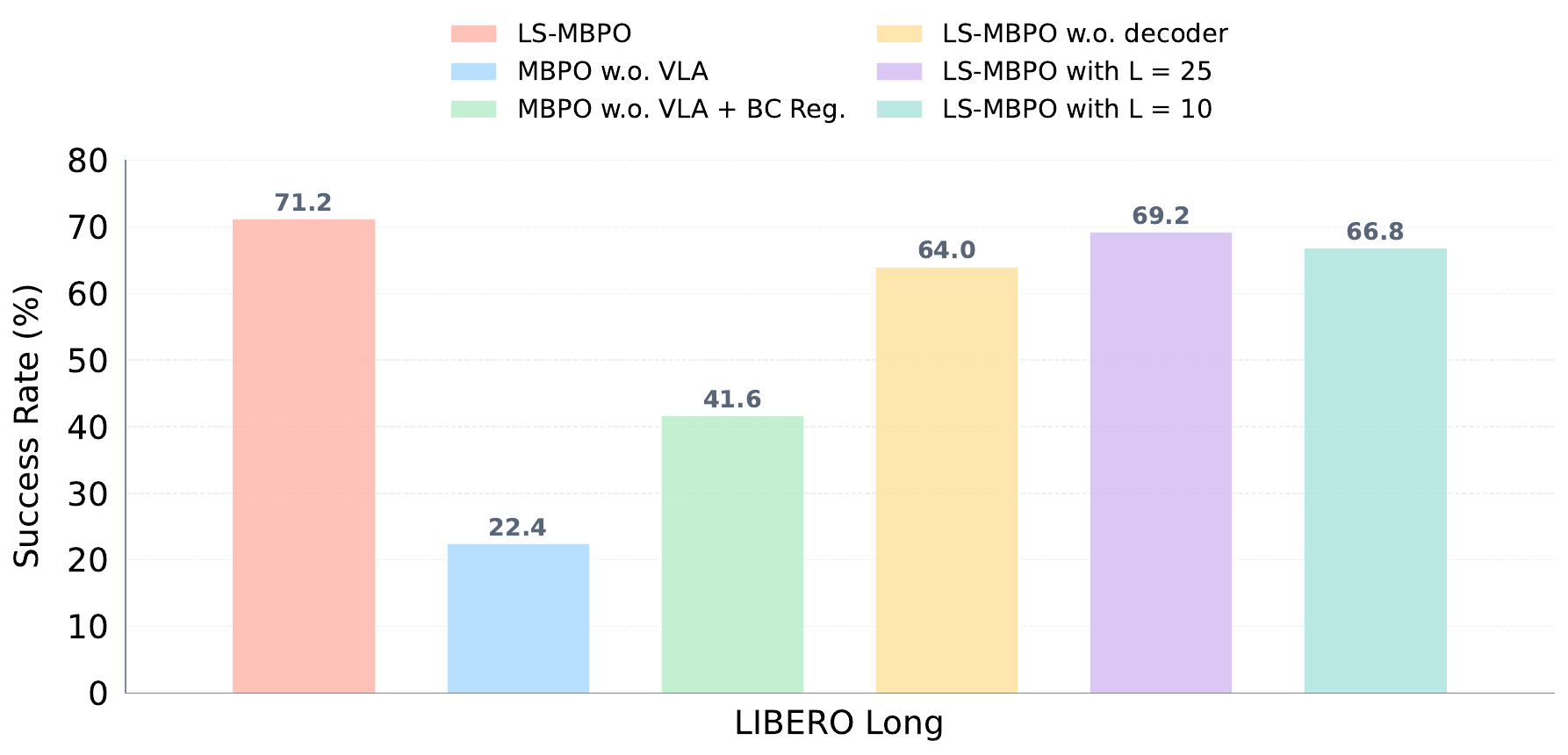}
    \vspace{-0.5em}
    \caption{Ablation results of \methodname~on LIBERO Long task.}
    \label{fig:appendix_libero_long_ablation}
    \vspace{-1.5em}
\end{figure}

To evaluate the impact of the decoder in latent world model training, we compare our world model with a variant trained without the decoder, where the image reconstruction loss in Eq.~\ref{eq:wm_loss} is removed. We then use both world models for subsequent MBRL with latent subchunk rollouts. The results in Fig.~\ref{fig:appendix_libero_long_ablation} show that our world model achieves greater policy improvement, suggesting that the reconstruction objective helps preserve additional information in latent representations that is beneficial for downstream policy learning.
It is observed in LeWorldModel~\cite{LeWM} that removing the decoder leads to better performance in planning. We think that this finding does not contradict our results, as planning and policy learning may have different representation requirements. Planning often benefits from compact and abstract representations that capture task-relevant dynamics while filtering out unnecessary visual details. In contrast, RL-based policy learning involves learning state-dependent action selection and value estimation, which may benefit from retaining richer information about the environment and object states.

\subsection{Ablation on Imagined Rollout Length}
We further conduct an ablation study on the imagined rollout length $L$, measured in environment steps. We fix the base action chunk length at $H=50$ and the subchunk size at $h=5$, and compare three settings: $L=10,25,50$, with $L=50$ used by default in \methodname.
The results in Fig.~\ref{fig:appendix_libero_long_ablation} show that longer imagined rollouts yield better policy performance within the evaluated range, suggesting that the learned world model provides useful training experience over these horizons.
A possible explanation is that longer rollouts under acceptable compounding errors are more likely to reach successful states from earlier branch starting points, providing informative binary success rewards for policy and value learning.

\subsection{Example Rollouts of Latent World Model}
\label{sec:example_rollouts}
\begin{figure*}[t]
    \centering
    \vspace{0em}
    \includegraphics[width=0.95\linewidth]{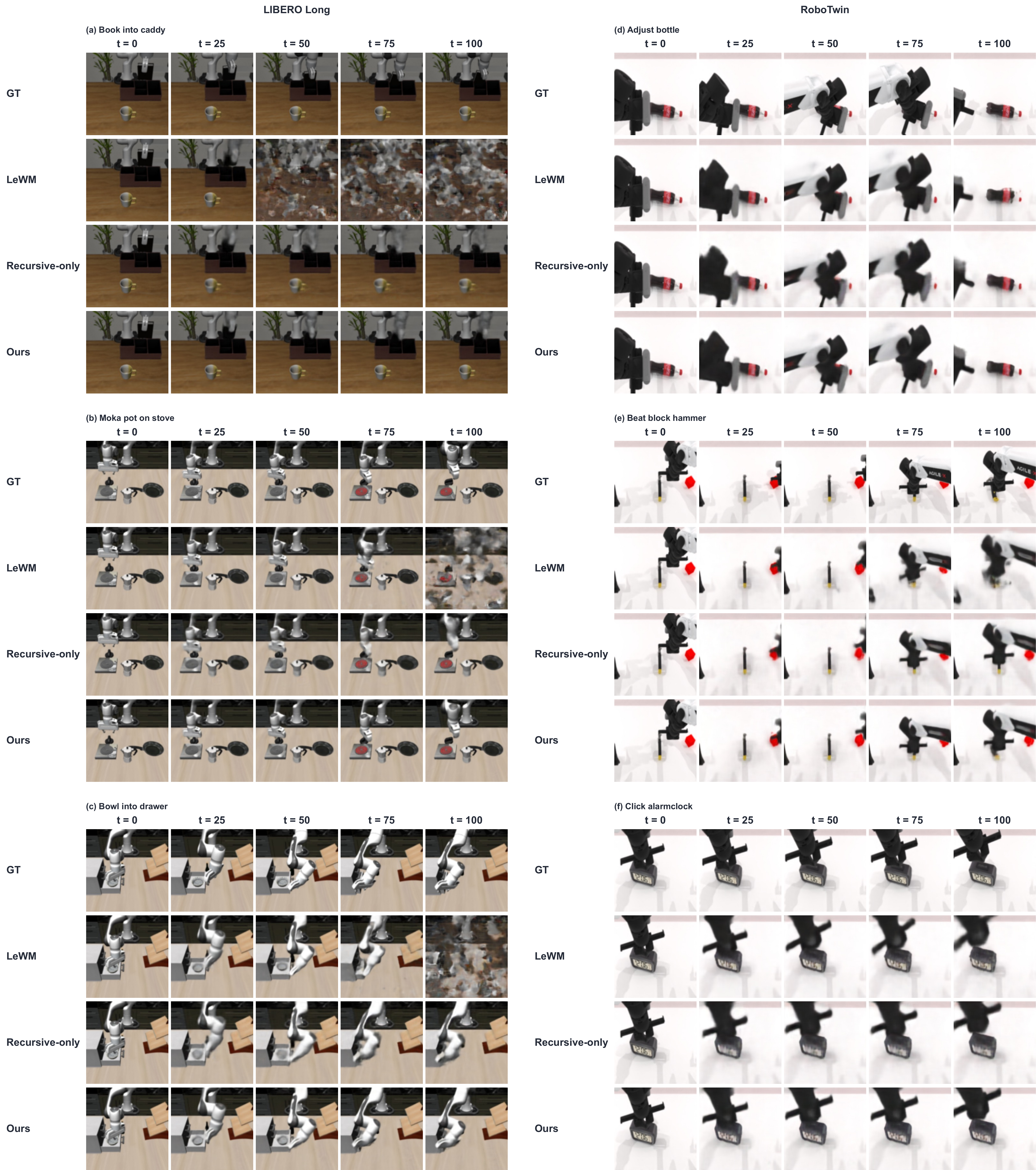}
    \vspace{-0em}
    \caption{Example rollouts of the latent world models in simulation tasks.}
    \label{fig:wm_rollout}
    \vspace{-0.0em}
\end{figure*}

In Fig.~\ref{fig:wm_rollout}, we visualize and compare the recursive rollouts of our latent world model trained with Hierarchical Forcing against a LeWM-style world model and a recursive-only world model described in Tab.\ref{tab:ablation_hierarchical_forcing}. We select three tasks from LIBERO Long and three tasks from RoboTwin, and sample one validation trajectory for each task. For each trajectory, we extract a 100-step segment and replay the corresponding action sequence through both world models.

We observe that the three models exhibit relatively similar reconstruction quality during the early stages of rollout. However, starting from $t=50$, the LeWM-style model begins to produce substantial deviations in several trajectories, particularly those from LIBERO Long. By $t=100$, its predictions become severely corrupted in all three LIBERO Long trajectories. 
The recursive-only world model does show better rollout quality compared to LeWM, but still falls short of our world model in some details of the image.
These qualitative results provide further evidence that Hierarchical Forcing effectively mitigates compounding errors during long-horizon recursive rollouts.

\begin{figure*}[t]
    \centering
    \vspace{0em}
    \includegraphics[width=0.95\linewidth]{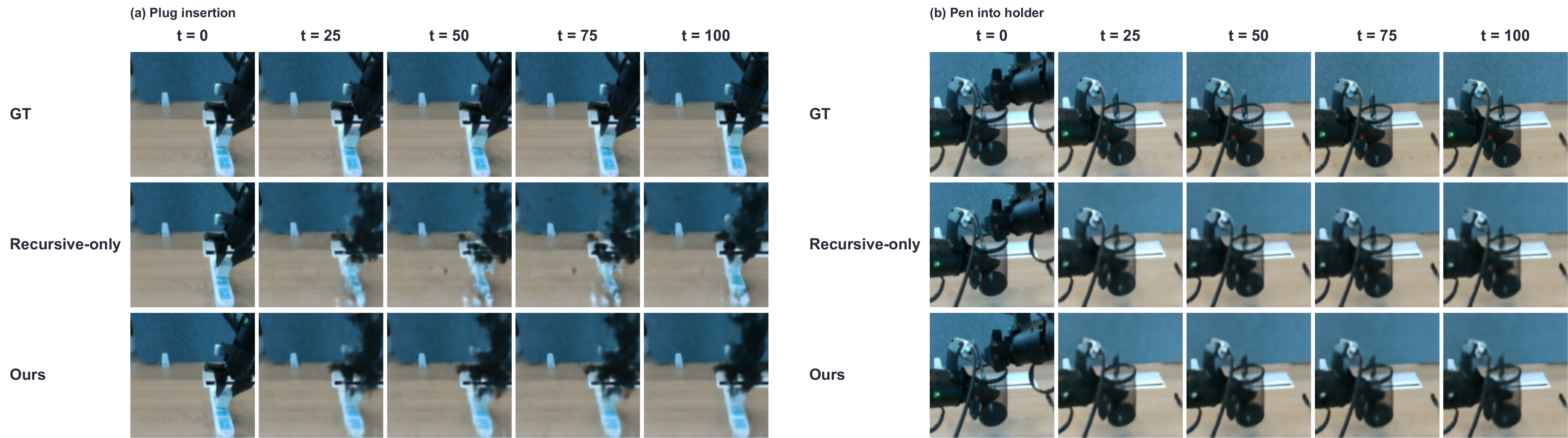}
    \vspace{-0em}
    \caption{Example rollouts of the latent world model in real-world tasks.}
    \label{fig:aloha_wm_rollout}
    \vspace{-0.0em}
\end{figure*}

We also provide the example rollouts of two real-world tasks of the learned latent world model in Fig.~\ref{fig:aloha_wm_rollout}.

\section{Algorithm Description}
The practical implementation of \methodname~is presented in the form of pseudo-code in Algorithm~\ref{alg:ls_mbpo}.

\begin{algorithm*}[t]
    \caption{\textbf{L}atent \textbf{S}ubchunk Rollout for \textbf{M}odel-\textbf{B}ased \textbf{P}olicy \textbf{O}ptimization (\methodname)}
    \textbf{Required}: an offline dataset $\mathcal{D}$, an expert dataset $\mathcal{G}$, a pretrained VLA $\pi_{\mathrm{VLA}}$\\
    \textbf{Output}: the optimized residual policy $\pi_\psi$.
    \begin{algorithmic}[1]
        \STATE Initialize the latent encoder $E_\theta$, dynamics predictor $F_\phi$, and CNN decoder.
        \STATE Initialize the residual policy $\pi_\psi$ and critic networks.
        \STATE // Training latent world model with Hierarchical Forcing
        \WHILE{world model training not converged}
            \STATE Sample trajectories from $\mathcal{D}$ and encode observations into latent representations.
            \STATE Train $F_\phi$ with Hierarchical Forcing using $\mathcal{L}_{\mathrm{pred}}$.
            \STATE Update $E_\theta$, $F_\phi$, and the decoder using Eq.~(\ref{eq:wm_loss}).
        \ENDWHILE
        \STATE // Training reward and progress predictors
        \STATE Construct positive and negative samples from $\mathcal{G}$ and $\mathcal{D}$.
        \STATE Train the success detector $\xi$ using binary success labels.
        \STATE Train the progress predictor $g_p$ using task progress supervision from $\mathcal{G}$.
        \STATE Freeze the latent world model $\mathcal{W}$, success detector $\xi$, progress predictor $g_p$, and pretrained VLA $\pi_{\mathrm{VLA}}$.
        \STATE // Training residual policy
        \WHILE{residual policy training not converged}
            \STATE Periodically collect real-world trajectories $\mathcal{K_{\mathrm{start}}}$ as branch starting points by rolling out with the current residual policy and VLA $\pi_{\mathrm{VLA}}$.
            \STATE Estimate task progress for states in $\mathcal{K_{\mathrm{start}}}$ using $g_p$.
            \STATE Construct the branch-start sampling distribution using PBC (Eq.~(\ref{eq:pbc_sampling})).
            \STATE Sample branch starting observations according to the progress-conditioned distribution.
            \FOR{each sampled branch starting observation $o_t$}
                \STATE Generate a base action chunk $B_t$ using the frozen VLA $\pi_{\mathrm{VLA}}(o_t,\ell)$.
                \STATE Divide $B_t$ into $N$ subchunks, each of length $h$.
                \STATE Initialize the latent history window $\widehat{\mathcal{H}}_0$ with the encoded observation.
                \FOR{$j=0,\ldots,N-1$}
                    \STATE Generate an action correction $\Delta A_j$ using $\pi_\psi(\hat{z}_{t+jh},B_t^{(j)},\ell)$.
                    \STATE Obtain the corrected action subchunk $\widetilde{A}_j$ using Eq.~(\ref{eq:action_refinement}).
                    \STATE Predict the next latent state using $F_\phi$ and the corrected action subchunk.
                    \STATE Compute the binary reward using the success detector $\xi$ (Eq.~(\ref{eq:success_reward})).
                    \STATE Store the imagined transition in the rollout buffer.
                \ENDFOR
            \ENDFOR
            \STATE Update $\pi_\psi$ and the critic networks using imagined transitions with TD3+BC.
        \ENDWHILE
        \STATE \textbf{return} the optimized residual policy $\pi_\psi$.
    \end{algorithmic}
    \label{alg:ls_mbpo}
\end{algorithm*}

\end{document}